# Turkish Sentiment Analysis Using Machine Learning Methods: Application on Online Food Order Site Reviews

Özlem AKTAŞ [1], Berkay COŞKUNER [1,*], İlker SONER [1]

[1] Dokuz Eylul University, Faculty of Engineering, Department of Computer Engineering, Turkey

**Abstract**
Satisfaction measurement, which emerges in every sector today, is a very important factor for many companies. In this study, it is aimed to reach the highest accuracy rate with various machine learning algorithms by using the data on Yemek Sepeti and variations of this data. The accuracy values of each algorithm were calculated together with the various natural language processing methods used. While calculating these accuracy values, the parameters of the algorithms used were tried to be optimized. The models trained in this study on labeled data can be used on unlabeled data and can give companies an idea in measuring customer satisfaction. It was observed that 3 different natural language processing methods applied resulted in approximately 5% accuracy increase in most of the developed models.

## 1. Introduction

Sentiment analysis is the process of analyzing and labeling the emotion created by the being that can create emotion. Today, when sentiment analysis is mentioned, studies on human and Twitter data usually come to mind. In the coming years, we may need to do sentiment analysis even from the human-like interfaces of the machines. However, the issue we are focusing on right now is a part that is closely related to both the academic world and the business world. Customer satisfaction is always at the forefront. Although this satisfaction is tried to be measured by various methods (questionnaire etc.), it is insufficient after a while and needs to be automated. This study enables us to analyze real emotion among the complex writings of people in an automated way.

To summarize in one sentence, sentiment analysis is the process of classifying the emotion that a person reveals together through the communication that the person uses, as positive or negative.

Every day, more data is created than human processing. If all people in the world quit all their work and try to read the data produced in just one day, it will not be successful. Therefore, there is a need for a more efficient way of processing data. There are so much data that cannot be detected with the naked eye. When you look at a table or reviews for a restaurant or comments for a product, you have a general idea, but you will never see the final result because you cannot read all of the comments. At this point, models are developed for processing the data. The purpose of this study is to help adapt the sentiment analysis method to real life and to test its accuracy.

Sentiment analysis studies can be performed on various predetermined emotions. It can also be diversified with other emotions such as fear, anger, sadness. In this study, it was evaluated only as positive and negative. The developed model shows whether a given sentence is positive or negative. While making this classification, user comments on Yemeksepeti.com site were used as data. The data set was not used ready, it was created.

The size of the created data set is approximately 676 thousand. Since the data set is completely homogeneous, no model tends to be positive or negative, there are 338 thousand positive and 338 thousand negative comments. The dataset was created using the Selenium module in Python.

There are 3 different techniques used in the study.
- Lemmatization
- Word Correction
- Keyboard (Our Method)

The Keyboard method is either used inside the Word Correction method or not. It cannot be used separately. There are 6 different datasets created using variations of these methods. These datasets were created to determine the effectiveness of the methods. All of the data sets are differentiated as 90% train 10% test randomly and have not been subjected to any processing other than the above-mentioned processes. The process of converting words to lowercase is common to all, as it is an operation done at the very beginning. The process of removing nonsense words and stop words is a common operation, as is converting words to lowercase. 6 different data sets used in the study are:

- Word Correction + Lemmatization (Default Dataset)
- Word Correction
- Lemmatization
- Word Correction without Keyboard
- Word Correction without Keyboard + Lemmatization
- No Operation

The difference in accuracy between the default dataset and the no-operation dataset will be based to see the final results of the study.

The major contributions of the study to the literature are:

- A dataset of homogeneous comments with a label of a size not previously available in Turkish.
- The potential contribution of the keyboard method to sentiment analysis and similar studies.
- Effective use of Lemmatization and Word Correction methods and observing their effects on similar studies.

To summarize the study, it was aimed to classify the comments as positive or negative by using the food basket comments. Datasets created with variations of 3 different methods and different machine learning algorithms were compared.

## 2. Previous Works

Similar studies conducted before us did not have such a large data set and keyboard method. Some similar studies are given in this chapter.

In the study of Erşahin et al. [1], a hybrid dictionary-based algorithm is used together with machine learning algorithms. The algorithms have been tested separately in 3 different data sets and the differences between them have been observed. The size of these data sets is 2000, 11000, and 50000. SVM, J48 and NB algorithms are used. They obtained the best result with an accuracy of 91% using SVM + SentiTurkNet library. Compared to single methods, hybrid methods show an average of 7% higher accuracy values.

In another study of Emekli et al. [2], it is aimed to classify Turkish tweets sent to the leading GSM operators of Turkey as positive and negative. In this study, deep learning methods are suggested for the classification of tweets. The posted tweets were first prepared with Natural Language Processing methods, and then they were classified and compared with deep learning methods such as Convolutional Neural Networks, Recurrent Neural Networks, and Long Short-Term Memory models. According to the results of the deep learning models, Convolutional Neural Networks could not increase the performance up to 9 epoch values on small data sets. Long Short-Term Memory, on the other hand, achieved a more successful result in the large datasets compared to other methods with a 98.64% accuracy rate. Recurrent Neural Networks achieved higher performance in the smaller datasets with a 98.8% success rate. It has been observed that the smaller the data set, the faster the performance in the Recurrent Neural Networks is learning.

In the study of Yilmaz et al. [3], offensive language was detected on the OffensEval dataset. This dataset consists of 31756 tweets. While 6131 of them have offensive content, 25625 of them consist of non-aggressive content. Accordingly, an untagged collection of approximately 1 million tweets was prepared. Afterward, the effect of word representations obtained from the labeled data in the OffensEval dataset, and the word representations obtained from the large untagged corpus on the classification performance were compared. Long Short-Term Memory (LSTM) and Bidirectional Long Short-Term Memory (BiLSTM) networks are used as machine learning models in the study. The classification performances of these deep neural networks are evaluated as accuracy, recall and precision, and F-score. While examining the effects of the extended corpus on deep learning models, it was determined that the success of the model was increased by using the expanded corpus. With this method, performance has been improved by approximately 40% to 47% compared to the F-score value, which is a measure of test accuracy. In addition, LSTM performed better than the deep learning models used. The performance values of the LSTM model were approximately 86% accuracy, 55% sensitivity, 68% precision, and 61% F-score, respectively. Performance values of the BiLSTM model were approximately 86% accuracy, 55% sensitivity, 66% precision, and 60% F-score, respectively.

A study of Baştürk [4] has been published on Kaggle with egebasturk1 account. A single-layer LSTM model was created with a total of 8570 Yemeksepeti comments. A wide variety of data preprocessing has been performed on the data by using Zemberek library, which is widely used in the NLP field. As a word representation, vectors prepared previously with word2vec method were used. 84% accuracy was taken as a result of the study. The project owner mentioned the insufficiency of word2vec vectors as self-criticism. In our eyes, a good accuracy value was obtained in this study because the structure of the neural network is weak, word vectors are weak, but data preprocessing is strong.

In the study of Yelmen [5], sentiment analysis was performed using various methods. These methods are Neural Network, Support Vector Machine, and Centroid Based Algorithm. In the study, 2 different methods were used, namely genetic algorithm, to find attributes. Accuracy, F-Measure, Recall, Precision values were used as performance criteria. In the data pre-processing phase, the root-finding process was performed using the Zemberek library.

The study of Aytuğ [6] on Turkish Tweets is aimed to have information about global events and to help crisis situations. The dataset contained 5300 positive and 5300 negative tweets with a total of 10600. Some natural language processing methods have been used together with various machine learning algorithms. Instead of the commonly used word2vec algorithm, the N-Gram algorithm is used. Accuracy, F-Measure, and AUC values were used among performance metrics. The highest success rate was achieved with the Bayes algorithm in the data set using the features obtained by using 1-Gram and 2-Gram common.

Another Turkish Sentiment Analysis on Turkish tweets but this study was about a specific topic. Albayrak et al. [7] did this study to find out what people's comments are about 'paid military service' which was one of the most popular topics of the period. 12739 tweets about the topic were collected to apply sentiment analysis. The messages obtained from Twitter were compared as a word with the words in the SentiTurk data set. The polarity score obtained for each tweet was evaluated according to their positivity, negativity, and neutral status. The tweets got a score according to the number of positive and negative words in their messages. In other words, no machine learning method was used in this study. Sentiment analysis was attempted by comparing the existing library and tweets.

In this research of Alpkoçak et al. [8], The outputs of different machine learning methods were compared. In this study, TREMO dataset was used. Sentiment analysis is considered as a text classification problem and a different approach than normal approaches has been tried to be used. A study was conducted on 6 different labeled emotions. These feelings are: happiness, fear, anger, sadness, disgust, and surprise, were used as emotion categories. In the study, root finding was done with F5, and TF-IDF was applied. As the performance metric, accuracy is used. The main purpose of the study was to develop a better model than the SVM method, one of the methods used in the past and considered a successful method. After experimenting with various layers and numbers of neurons, an artificial neural network was developed with accuracy (86%) 0.0045% greater than SVM.

The study of İlhan et al. [9] is another example of Sentiment Analysis on Turkish tweets and it focuses mainly on sentiment analysis of Twitter which is one of the popular social media sites used by numerous users where users publish a status update in the form of tweets. The dataset that Nagehan İlhan and Duygu Sağaltıcı used has 1,578,627 classified tweets. And the model was built for classifying tested tweets into positive and negative sentiment. To perform this analysis, an intelligent model has been created by using machine learning methods such as Naïve Bayes and Support Vector Machine and the compared results has been given. In conclusion, Unigram and Bigram Support Vector Machine give the best results with 64% accuracy. In the total dataset, Naive Bayes gave a 42% success rate, while VADER had a 27% success rate. When looking at the results, it is seen that the results are improved when Unigram and Bigram are used together with Support Vector Machine.

In the study Sarıman et al. [10] of collected the tweets that have been posted about coronavirus in Turkey since 11 March 2020, In this study using logistic regression, approximately 2 million tweets were used. The samples were collected on 5 different topics, but they were divided into 2 parts as positive and negative while classifying.

Since success rates were not at the desired level according to randomly generated training sets at first, training sets were determined according to positive and negative word groups. As a result of this change, AUC values increased from an average of 0.80 to an average of 0.95. And when the general results are examined, it is observed that people's comment on government's mask application is generally evaluated as positive, but other applications are generally evaluated as negative.

Offensive language was detected on the OffensEval dataset in the study of Yilmaz et al. [11]. This dataset consists of 31756 tweets. While 6131 of them have offensive content, 25625 of them consist of non-aggressive content. Accordingly, an untagged collection of approximately 1 million tweets was prepared. Afterward, the effect of word representations obtained from the labeled data in the OffensEval dataset, and the word representations obtained from the large untagged corpus on the classification performance were compared. Long Short-Term Memory (LSTM) and Bidirectional Long Short-Term Memory (BiLSTM) networks are used as machine learning models in the study. The classification performances of these deep neural networks are evaluated as accuracy, recall and precision, and F-score. While examining the effects of the extended corpus on deep learning models, it was determined that the success of the model was increased by using the expanded corpus. With this method, performance has been improved by approximately 40% to 47% compared to the F-score value, which is a measure of test accuracy. In addition, LSTM performed better than the deep learning models used. The performance values of the LSTM model were approximately 86% accuracy, 55% sensitivity, 68% precision, and 61% F-score, respectively. Performance values of the BiLSTM model were approximately 86% accuracy, 55% sensitivity, 66% precision, and 60% F-score, respectively.

In the study of Gezici et al. [12] movie reviews are used. In addition to supervised methods, a lexicon-based method was also used. In addition, the Turkish polarity lexicon called SentiTurk was also used.

The comments are taken from the movie site called the big screen. The best 2 results are from Naive Bayes and Support Vector Machine algorithms. They have found that working with large lexicons always gives better results. The dataset is homogeneous and includes 5300 positive and negative comments. When all features are used, approximately 75% success was achieved in both algorithms.

## 3. Methodologies

The first step is to prepare the raw data set. This unprocessed data set was obtained from Yemek Sepeti via Selenium module in Python. Selenium module is a module for getting information from websites using Python. By taking HTML/CSS codes and making them more suitable for the eye, it provides the desired data from websites.

After the raw form of the dataset is prepared, its variations should be created according to the methods. Word2vec [13] (Mikolov et al., 2013) models have been developed to represent words specifically for all datasets. General flow of the project is given in Figure 1.

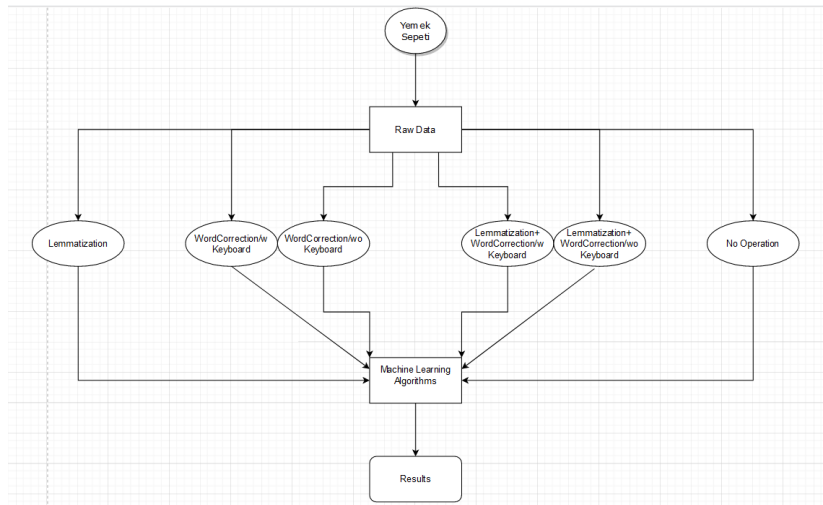

**Figure 1.** *Flow of Study.*

Lemmatization is the process of converting a word to its base form. It helps to provide an increase in accuracy by putting words that have the same meaning but are spelled differently in the data set into the same form. In Figure 2 and Figure 3 it can be observed that the lemmatization library works quite well and helps to obtain better and more accurate results with the studied datasets. The library used for lemmatization is called zemberek-python. The upperparts are the raw data, the lower parts are lemmatized using the zemberek-python library.

```
['Yazılan', 'notları', 'dikkate', 'almadığınız', 'için', 'size', 'şöyle', 'kötü', 'bir', 'puan', 'verelim']
['yazmak', 'not', 'dikkat', 'almamak', 'için', 'siz', 'şöyle', 'kötü', 'bir', 'puan', 'vermek']
```

**Figure 2.** *Example Lemmatization 1.*

```
['Tam', 'bir', 'buçuk', 'saatte', 'geldi', 'Defalarca', 'aramamıza', 'rağmen', 'telefonu', 'asla', 'açmadılar']
['tam', 'bir', 'buçuk', 'saat', 'gelmek', 'defalarca', 'aramak', 'rağmen', 'telefonu', 'asla', 'açmamak']
```

**Figure 3.** *Example Lemmatization 2.*

The library used for the word correction method is called trnlp. This library corrects Turkish misspelled and incorrect words. It also converts words written in English characters into Turkish characters. These two processes are important for the accuracy increase and for training models efficiently.

```
['Surekli', 'icerik', 'yanlis', 'geliyor']
['sürekli', 'içerik', 'yanlış', 'geliyor']
```

**Figure 4.** *Example of Word Correction.*

The working principle of the created keyboard method is based on a simple principle: the human factor. Most spelling mistakes made on phones or computers actually occur around the letter you want to type. The Keyboard method tries to calculate which of these probabilities is greater and returns you the best possible result. This method is used together with the Word Correction method. The Word Correction method suggests 10 words for a word correction. The Keyboard algorithm, on the other hand, calculates between the first 2 suggested words and the originally written word and tries to find out which of the 2 suggestions is more likely. Keyboard matrix created as seen in Figure 5.

```
a z s w q
b v g h n
c x d f v
ç ş l k m
d e r f c x s
e w s d r
f r t g v c d
g t y h b v f
ğ p ş i ü
h y u j n b g
ı u j k o
i ü ğ ş
j u ı k m n h
k ı o l ö m j
l o p ş ç ö k
```

**Figure 5.** *Part of Keyboard Matrix.*

As it can be seen in Fig.5, there is an alphabetical order in the first column. In the other columns, there are the letters around the current letter in the Turkish Q keyboard.

### 3.1. Parameter Tuning

The neural network model was developed using TensorFlow, and all other models were developed using the sklearn module. Each algorithm has its own parameters and the best parameters need to be found to make the algorithm work well. Some algorithms have a low number of parameters or the ratio of these parameters to the accuracy value, but there are algorithms that have the opposite. The neural network parameter optimization, which is the first of the methods used, was systematically performed manually on the default data set. The values seen in Table 1 include the results of the experiments carried out with a high number of neurons and hidden layers, without taking into account the runtimes.

**Table 1.** *Neural Network Optimization 1.*

| Batch Size | Epoch | Structure | Neuron Type | Accuracy |
|---|---|---|---|---|
| 1024 | 5 | 64-64-16-8-4-1 | GRU | 0,8389 |
| 2048 | 5 | 64-64-16-8-4-1 | GRU | 0,8257 |
| 512 | 5 | 64-64-16-8-4-1 | GRU | 0,8515 |
| 256 | 5 | 64-64-16-8-4-1 | GRU | 0,8606 |
| 128 | 5 | 64-64-16-8-4-1 | GRU | 0,8618 |
| 2048 | 5 | 128-64-16-8-4-1 | GRU | 0,8384 |
| 2048 | 5 | 512-64-16-8-4-1 | GRU | 0,8354 |
| 2048 | 1 | 64-64-16-8-4-1 | GRU | 0,8029 |
| 2048 | 3 | 64-64-16-8-4-1 | GRU | 0,8207 |
| 2048 | 7 | 64-64-16-8-4-1 | GRU | 0,8212 |
| 2048 | 5 | 512-128-16-8-4-1 | GRU | 0,8443 |
| 2048 | 10 | 64-64-16-8-4-1 | GRU | 0,8400 |
| 2048 | 12 | 64-64-16-8-4-1 | GRU | 0,8410 |
| 2048 | 5 | 512-256-128-16-8-4-1 | GRU | 0,8102 |
| 2048 | 17 | 64-64-16-8-4-1 | GRU | 0,8518 |
| 2048 | 20 | 64-64-16-8-4-1 | GRU | 0,8490 |
| 2048 | 5 | 64-64-64-64-64-64-1 | GRU | 0,8449 |
| 2048 | 5 | 64-64-16-8-4-1 | CuDNNGRU | 0,8317 |
| 2048 | 25 | 64-64-16-8-4-1 | CuDNNGRU | 0,8552 |
| 2048 | 5 | 64-64-16-8-4-1 | CuDNNGRU | 0,8358 |
| 2048 | 5 | 64-128-128-128-128-128-1 | GRU | 0,8555 |
| 2048 | 5 | 64-256-256-256-256-256-1 | GRU | 0,8600 |
| 2048 | 5 | 64-64-16-8-4-1 | CuDNNLSTM | 0,8143 |
| 2048 | 5 | 64-256-256-256-256-256-256-1 | CuDNNGRU | 0,8544 |
| 2048 | 5 | 64-64-16-8-4-1 | Bidirectional CuDNNGRU | 0,8546 |

As given in Table 2, experiments were performed with smaller numbers of neurons and hidden layers, and operating times were taken into account.

Here are the best values obtained after manual work.

- Batch Size = 32
- Layer Type = Bidirectional CuDNNGRU
- Epoch = 10
- Structure = 8-8-8-1

Table 2. *Neural Network Optimization 2.*

| Batch Size | Epoch | Structure | Neuron Type | Runtime(s) | Accuracy |
|---|---|---|---|---|---|
| 32 | 10 | 8-8-8-8-8-1 | GRU | 2500 | 0,8569 |
| **32** | **10** | **8-8-8-8-8-1** | **Bidirectional CuDNNGRU** | **800** | **0,8612** |
| 2048 | 10 | 8-8-8-8-8-1 | Bidirectional CuDNNGRU | 16 | 0,8492 |
| 2048 | 20 | 8-8-8-8-8-1 | Bidirectional CuDNNGRU | 16 | 0,8552 |
| 2048 | 40 | 8-8-8-8-8-1 | Bidirectional CuDNNGRU | 16 | 0,8578 |
| 8 | 5 | 8-8-8-8-8-1 | Bidirectional CuDNNGRU | 1250 | 0,8595 |
| 32 | 10 | 8-8-8-1 | Bidirectional CuDNNGRU | 215 | 0,8637 |
| 32 | 10 | 8-8-1 | Bidirectional CuDNNGRU | 130 | 0,8517 |
| 32 | 10 | 64-8-8-1 | Bidirectional CuDNNGRU | 215 | 0,8601 |
| 32 | 10 | 64-8-1 | Bidirectional CuDNNGRU | 135 | 0,8509 |
| 512 | 10 | 8-8-8-8-8-1 | Bidirectional CuDNNGRU | 27 | 0,8585 |
| 512 | 10 | 64-8-1 | Bidirectional CuDNNGRU | 13 | 0,8524 |
| 8 | 10 | 8-8-8-1 | Bidirectional CuDNNGRU | 714 | 0,8590 |
| 32 | 10 | 8-8-8-1 | Bidirectional CuDNNLSTM | 1065 | 0,8576 |
| 32 | 10 | 8-8-8-1 | CuDNNGRU | 440 | 0,8558 |
| 32 | 10 | 8-8-8-1 | CuDNNLSTM | 444 | 0,8499 |

As 2 experiments show, in Table 1 and Table 2, small-batch and optimum epoch number are effective in giving maximum performance. In addition, it is obvious that there is no proportionality between the complexity of the model and the accuracy value.

A systematic study could not be performed in the support vector machine and the k-nearest neighbor method due to the long run times. The best values from several random trials:

Support Vector Machine

- Kernel = Polynomial
- C = 0.1
- Gamma = 0.1

K-Nearest Neighbor

- K = 7
- Algorithm = Auto
- Weights = Uniform

The GridSearchCV function, which is a function of the sklearn module, was used in Naive Bayes and Linear Regression algorithms. The values are as follows:

Naïve Bayes

- var_smoothing = 0.151

Linear Regression

- fit_intercept = True
- normalize = True

The studies were continued with the same parameters in all data sets.

### 3.2. Evaluation Metrics

Accuracy, f-measure, precision, and recall were used as calculation metrics. In addition to these, the Mean Squared Error (MSE) metric is also used to measure performance between algorithms. However, for the simplicity of the tables, only the accuracy and MSE values are included. Since linear regression outputs are continuous data, methods such as MSE are used to measure the success of the algorithm. Therefore, there are no accuracy values for linear regression in the tables.

- True Positive (TP), prediction is positive while real value is positive.

- True Negative (TN), prediction is negative while real value is negative.

- False Positive (FP), prediction is negative while real value is positive.

- False Negative (FN), prediction is positive while real value is negative.

$$Accuracy = \frac{TP+TN}{TP+TN+FP+FN} \quad (1)$$

$$Precision = \frac{TP}{TP+FP} \quad (2)$$

$$Recall = \frac{TP}{TP+FN} \quad (3)$$

$$Fm = 2 * \frac{(Precision*Recall)}{(Precision+Recall)} \quad (4)$$

MSE(Mean Squared Error) is the error metric that tells how close a regression line is to the predictions. It is both positive and usually greater than 0 because it is derived from the square of the Euclidean distance. It was used to measure the success of the regression model.

$$MSE = \frac{1}{n}\sum_{i=1}^{n}(Yi - Y'i)^2 \quad (5)$$

## 4. Results

In this study, as seen in Table 9, there is an accuracy value difference of approximately 5% between the unapplied state of all methods and the applied state of all methods. Except for the k-nearest neighbor algorithm, there is a noticeable linear difference in the other algorithms. In the case of lemmatization, word correction, and keyboard methods applied separately, approximately 1% increase in accuracy value was observed for each. The results on the default data set can be seen in Table 3.

**Table 3.** *Default Data Set Results.*

| Default Data Set | Accuracy | MSE | Runtime(s) |
|---|---|---|---|
| Neural Network | %86,37 | 0,136 | 2150 |
| Support Vector Machine | %87,24 | 0,127 | 21977 |
| K-Nearest Neighbor | %81,88 | 0,181 | 907 |
| Naive Bayes | %83,55 | 0,164 | 175 |
| Linear Regression | - | 0,131 | 69 |

In the following tables, the results of the remaining 5 data sets can be observed.

**Table 4.** *Word Correction Data Set Results.*

| Word Correction Data Set | Accuracy | MSE | Runtime(s) |
|---|---|---|---|
| Neural Network | %83,9 | 0,135 | 2152 |
| K-Nearest Neighbor | %83,2 | 0,133 | 903 |
| Naive Bayes | %82,05 | 0,131 | 177 |
| Linear Regression | - | 0,134 | 72 |

**Table 5.** *Lemmatization Data Set Results.*

| Lemmatization Data Set | Accuracy | MSE | Runtime(s) |
|---|---|---|---|
| Neural Network | %84,06 | 0,138 | 2137 |
| K-Nearest Neighbor | %83,8 | 0,133 | 885 |
| Naive Bayes | %82,14 | 0,142 | 165 |
| Linear Regression | - | 0,132 | 71 |

**Table 6.** *Word Correction without Keyboard Data Set Results.*

| Word Correction without Keyboard Data Set | Accuracy | MSE | Runtime(s) |
|---|---|---|---|
| Neural Network | %82,66 | 0,141 | 2002 |
| K-Nearest Neighbor | %84,1 | 0,134 | 1003 |
| Naive Bayes | %81,16 | 0,142 | 189 |
| Linear Regression | - | 0,140 | 71 |

**Table 7.** *Word Correction without Keyboard + Lemmatization Data Set Results.*

| Word Correction without Keyboard + Lemmatization Data Set | Accuracy | MSE | Runtime(s) |
|---|---|---|---|
| Neural Network | %84,94 | 0,140 | 1987 |
| K-Nearest Neighbor | %84,9 | 0,130 | 1132 |
| Naive Bayes | %82,98 | 0,135 | 167 |
| Linear Regression | - | 0,131 | 70 |

**Table 8.** *No Operation Data Set Results.*

| No Operation Data Set | Accuracy | MSE | Runtime(s) |
|---|---|---|---|
| Neural Network | %81,15 | 0,142 | 2003 |
| K-Nearest Neighbor | %83,2 | 0,131 | 912 |
| Naive Bayes | %80,34 | 0,135 | 154 |
| Linear Regression | - | 0,143 | 55 |

The table containing all the most recent results of the study is shown below. It has been observed that there is an increase of 1% for each of the methods applied separately, and 5% when applied together.

**Table 9.** *Final Results Accuracy/MSE.*

| Dataset/Algorithm | Neural Network | Naïve Bayes | K-Nearest Neighbor | Linear Regression |
|---|---|---|---|---|
| Default | %86,37/0,136 | %83,55/0,164 | %81,88/0,181 | -/0,131 |
| WordCorrector | %83,9/0,135 | %82,05/0,131 | %83,2/0,133 | -/0,134 |
| Lemmatizing | %84,06/0,138 | %82,14/0,142 | %83,8/0,133 | -/0,132 |
| WordCorrector w/o Keyboard | %82,66/0,141 | %81,16/0,142 | %84,1/0,134 | -/0,140 |
| WordCorrector w/o Keyboard + Lemmatizing | %84,94/0,140 | %82,98/0,135 | %84,9/0,130 | -/0,131 |
| No Operation | %81,15/0,142 | %80,34/0,135 | %83,2/0,131 | -/0,143 |

## 5. Conclusion

In this study, various machine learning methods were tested and tried to be improved using approximately 676 thousand comments. The adaptation of NLP applications to Turkish and their effects on accuracy values were observed. As seen in previous studies, it is a fact that the Support Vector Machine algorithm performs very well in NLP applications in any way. However, if there is a time constraint, it may be more useful to use other simple but effective algorithms. The main focus of the study was Artificial Neural Networks. The parameters were systematically optimized without using GridSearchCV. The accuracy value obtained was close to the Support Vector Machine algorithm. It is possible to say that the 2 algorithms go head-to-head, but when the subject is considered as a hyperparameter, it seems that artificial neural networks can be developed further. Likewise, there are parameters that can be optimized in the Support Vector Machine algorithm. Much better accuracy can be achieved when a little more time and computational power is added to the study. This study, unlike other studies, showed the effect of the Keyboard method in the field of Turkish NLP. In addition, a large data set and different machine learning algorithms have revealed the effect of different natural language processing methods.

Since such a large Turkish data set was not encountered in literature research, the data set was created from scratch. The generated dataset is published in Kaggle [14] to be used for everyone works and interested in this area. It is hoped that this data set and the results of the studies will contribute to future Turkish studies.

**Declaration of Interest**

As authors, we declare that we have no conflict of interest with anyone related to our work.